\documentclass[authoryear,preprint,review,12pt]{elsarticle}

\usepackage{epsfig}

\usepackage{amssymb}

\usepackage{tipa}
\usepackage{booktabs}
\newcommand{\sizefigs}{45mm}

\journal{Physica A: Stat.~Mech.~and its Applications}

\begin{document}

\begin{frontmatter}




\title{Deciding the status of controversial phonemes
using frequency distributions; \\
an application to semiconsonants in Spanish} 


\author{Manuel Ortega-Rodr\'{\i}guez\footnote{Corresponding author. \\
Email addresses: manuel.ortega@ucr.ac.cr (M.~Ortega-Rodr\'{\i}guez), hugo.solis@ucr.ac.cr (H.~Sol{\'\i}s-S\'anchez),
ricardo.gamboa@iftucr.org (R.~Gamboa-Alfaro).}, 
Hugo Sol{\'\i}s-S\'anchez, 
Ricardo Gamboa-Alfaro}  

\address{Escuela de F\'{\i}sica, 
Universidad de Costa Rica, 11501-2060 San Jos\'e, Costa Rica}

\begin{abstract}
Exploiting the fact that natural languages are complex systems,
the present exploratory article proposes a direct method based on 
frequency distributions that may be useful when making a
decision on the status of problematic phonemes, 
an open problem in linguistics.
The main notion is that
natural languages, which can be considered from a complex outlook 
as information processing machines, and which somehow
manage to set appropriate levels of redundancy,
already ``made the choice'' whether a linguistic unit is a phoneme or not,
and this would be reflected in a greater smoothness in a frequency versus rank graph.
For the particular case we chose to study, we 
conclude that it is reasonable to consider the Spanish
semiconsonant /w/
as a separate phoneme from its vowel counterpart /u/, on the one hand,
and possibly also 
the semiconsonant /j/ as a separate phoneme from its vowel counterpart /i/,
on the other.   
As language has been so central a topic in the study
of complexity, this discussion grants us, in addition, an
opportunity to gain insight into emerging properties in
the broader complex systems debate.
\end{abstract}


\end{frontmatter}


\section{Introduction}
\label{}

Natural languages are complex systems and as such are
expected to share many features with other complex systems [1,2].
The purpose of this article is to use this idea to present a direct
method
based on frequency distributions that may be useful 
in general when making a decision
on the status of problematic phonemes in natural languages,
an open problem in linguistics [3].
The potential usefulness becomes evident when comparing 
the simplicity of the method 
with the intricate nature of theories dealing with phoneme status
determination. 

The whole approach is based on the idea that natural languages
are complex systems and as such
have already ``made the decision'' on whether a particular
linguistic unit is a phoneme or not.

Complex systems manifest different types of emergent structures.  
Interestingly, the emergent structure is often 
similar across different types of physical systems and can be usually modeled as a network, language not being the exception [4]. 
Such networks usually have the property of being ``small-world''  
(i.e. possessing high connectivity in spite of not being chaotic), 
at least in some scales [5]. If not having a Zipf or Zipf-like 
distribution [6,7,8], the behavior
of the relevant parameters 
will at the very least show some degree of smoothness. 
The reason behind this type of distribution, it has been argued, 
is that it makes communication
efficient [9]. 
We use the notion of smoothness to suggest
an elegant criterion for phonemeness.

Natural languages have been shown to possess simple frequency versus rank
distributions at the word frequency level, where,
as we mentioned, 
Zipfian and near-Zipfian distributions
have been documented (Zipf
distributions were actually first observed in natural languages [10]).  
However, there is nothing particularly special about analyzing language at the 
level of words (which may not even be a precisely defined psychological class).
Such a particularity would actually be contrary to the spirit of complexity, 
as it would single out a particular scale.
In reality, sentence length also shows   
an emergent structure and smooth dependence of the parameters [8]. 
 
The appreciation of structure at the phoneme level, however, is 
somewhat hindered
by the fact that one has direct computational
access only to the letters and not easily to the actual phonemes, a 
situation complicated by the idiosyncratic nature of the spelling in 
some languages.\footnote{This might explain why phoneme distribution regularities 
have been much less studied than those of other linguistic units.}
This is why
the use of Spanish is useful. Although there is nothing special about 
the phonology of Spanish
language per se, its relatively phonemic orthography (including
diacritical marks) makes it suitable
for the kind of study this paper undertakes.
 
The method is thus immediately applied to the case study of
semiconsonants in Spanish.
We find the method to be clean and
obtain support for the claim that 
at least one of the Spanish semiconsonants is a separate phoneme, 
thus contributing decisively to 
a long-standing issue. (Deep, ``internal'' 
model mechanisms, however, are of course not directly revealed by the method,
at least at this stage.)

As we describe below, this exploratory method is coarse-grained,
as some important assumptions are made. For example, 
we average over important dialectal differences of Spanish.
As with other approaches to complex systems (see e.g. Ref.~[11]), a
``low resolution'' stance avoids the distraction
arising from details that are not essential to the phenomenon in question.
In particular, the approach does not go beyond a very rough
phonemization of the data.
The intentional avoidance of details also has the virtue of removing
possible biases.
There is no obvious reason why 
in principle
the method could not be applied to any phoneme candidate in any language, provided one has at least a rough phonemic transcription of a sample.

The present research has been done in the style of physics,
not linguistics, as the main guidelines were
simplicity, ``low resolution''  
and randomness (as in the selection of the sample text).  
Priority was given to a ``top down'' approach. 

We thus took a complicated, long-standing problem and used
a simple insight to solve it.
Detailed values of the entropy and redundancy,
although calculated, were actually not needed to establish the
main conclusions, as visual inspection of the graphs turned out to 
be enough, in a similar fashion as when visually identifying outliers 
in general.  
We consider this fact a strength of the approach. 
Physics has plenty of examples were insightful counting was the way out of  
baffling dead ends, as in the 
Gibbs paradox and the anomalous Zeeman effect.

\section{Method and case study: semiconsonants in Spanish}
\label{}
 
We expound here the method by describing it while we apply it directly to a case study,
namely the determination of the phonological status of semiconsonants
in Spanish (not there being to our knowledge anything particularly unique
about Spanish or about semiconsonants).
 
Establishing the phonological structure of a language is no easy task.
In particular, the phonemic status of semiconsonants in Spanish has been termed
``problematic,'' 
and there exists a long-standing discussion in the literature 
[12,13] 
(and Spanish is by no means unique in this sense).
As the problem is entangled with prosody and morphological features,
different competing attempts for a solution have been proposed, 
none of which resulting triumphant.
A different, ``third party'' approach to this problem,
especially one which is independent of traditional linguistic models,
is therefore 
timely and valuable.
 
The issue at hand is the following:
are the semiconsonant glides in words such as ``tiene'' and 
``bueno'' (represented by the letters ``i'' and ``u'', respectively) 
separate phonemes, or are they
merely allophones (i.e. phonetic variants) of the 
respective vowels /i/ and /u/?\footnote{Unless otherwise stated, 
we follow the conventions of the International Phonetic Alphabet.
We use slashes for phonemes and square brackets for sounds.}

Most linguists prefer to 
keep the ``economical'' interpretation, i.e. deny phoneme status
to semiconsonants. This is sensible because semiconsonant glides 
appear to be 
in complementary distribution with the respective related closed vowels
[i] and [u], 
except for a few inconvenient anomalies.\footnote{Hualde [15] offers the example
``pie'' [pie] (``I chirped'') versus ``pie'' [pje] (``foot'').} 
The main problem with this interpretation
is that it requires a priori knowledge of stress position,
and thus the introduction of stress as a phoneme.
Phonemic stress, however, 
has not been a historically preferred approach among linguists and
runs contrary to neurocognitive evidence [14].
The other option, namely granting phoneme status to semiconsonants,
is not perfect either, unfortunately, because it turns out that stress
does not become completely predictable even in this case [12], 
although it is true that it becomes much more predictable than in the 
economical case.
As the situation remains unresolved, 
frustration surfaces in the form of ``quasi-phonemic contrast'' 
discussions [15].  

We intend to solve this issue by examining carefully various 
frequency versus rank distributions
with different working hypotheses concerning phonemic candidates, 
and assessing possibilities by looking
at the behavior of the graphs, in particular, their smoothness.

Smoothness as a feature commonly appears in optimization processes.
Given that natural languages can be regarded as complex networks
showing universality and optimization 
(for example, in their small world property), 
we conclude that it makes sense to employ 
smoothness as a candidate marker for correct phoneme counting.
Moreover, the suitability of smoothness becomes clearer after the following
consideration.

As smoothness is a relative quality, it is important to establish
a benchmark for non-smoothness.
One of such benchmarks can be obtained, 
for example, by (artificially) treating /t/ and /d/ in Spanish as the 
same phoneme, even though it is known they are not. 
In such a case, the results (in the form of broken, jagged plots) 
look similar to what happens when one
takes /u/ and /w/ to be the same phoneme. 
Another benchmark for non-smoothness can be created by using 
letters (which in their obsolescence do not keep up with the language)  
instead of phonemes, as discussed below. 
Here again, the resulting jagged pattern resembles that of treating /t/ and /d/,
or /u/ and /w/, as the same phoneme. 
 
As discussed in the previous section, 
we use a coarse-grained method and some important assumptions are made.
For example, we have assumed 
that text is a good enough proxy for speech for the purposes of
our exploratory study. 
(A complete list of assumptions can be found in Appendix A.)

A random
text was chosen, its size being around 23 thousand characters (not counting spaces). 
After all characters were lower-cased and 
diacritical marks eliminated (see, however, below), a very simple phonemization
was carried out (as described in Appendix B), 
focusing on the most basic features of spelling conventions.
Frequency distributions for the characters were then calculated, 
ignoring spaces, punctuation marks, and non-alphabetic characters.
Smoothness and information theoretical values were finally computed.
We defined $F(n)$ as the fractional frequency of the $n$-th element
(putative phoneme), where $n$ runs from 1 to $N$, the total number of phonemes,
and
$D(n) \equiv F(n+1) - F(n)$.
A reasonable value for the smoothness [16] of a given frequency distribution was 
then calculated using the standard deviation of $D(n)$ and its mean:
\begin{equation}
  1/{\rm smoothness} \equiv \frac{{\rm SD}(D)}{|{\rm mean}(D)|} \, ,
\end{equation}
while the Shannon information per symbol [17] was computed using 
\begin{equation}
  H \equiv - \sum F(n) \log_2 F(n) \, ,
\end{equation}
where the sum runs from 1 to $N$. Redundancy is given by 
\begin{equation}
  R \equiv 1 - \frac{H}{\log_2 N} \, .
\end{equation}

\section{Results and discussion}
\label{}
 
The results are presented in Figs.~1 to 7 and in Table 1.
\begin{table}[htbp]
   \centering
\resizebox{\textwidth}{!}{
     \begin{tabular}{@{} lccccc @{}} 
      \toprule
           & Figure
           & $N$
           & Smoothness
           & Redundancy
           & Shannon  \\
           & 
           & 
           & 
           & 
           & information   \\
      \midrule
         Reference case (simple phonemization, Appendix B)  
         & 1 & 22 & 0.9214 & 12.11 \% & 3.92   \\
         Raw data (no phonemization)  
         & 2 & 27 & 0.7175 & 14.79 \% & 4.05  \\
         Semiconsonant /w/ as separate phoneme from /u/
         & 3 & 23 & 0.9635 & 12.74 \% & 3.95  \\
         Both semiconsonants /w/ and /j/ as phonemes  
         & 4 & 24 & 0.8683 & 12.18 \% & 4.03  \\
         As Fig.~4, but using diacritics to discern true diphthongs   
         & 5 & 24 & 0.8949 & 12.27 \% & 4.02  \\
         All diphthongs as phonemes  
         & 6 & 46 & 0.6007 & 20.08 \% & 4.41  \\
         Diphthongs containing /w/ or /j/ as phonemes 
         & 7 & 37 & 0.6418 & 19.73 \% & 4.18 \\
      \bottomrule
   \end{tabular}}
   \caption{Numerical data corresponding to the figures. 
            $N$ is the number of putative phonemes, whereas
            smoothness, redundancy and Shannon information (in bits per symbol)
            are computed using formulas (1), (3), (2), respectively.}
   \label{table2}
\end{table}
It is remarkable that, in spite of the usefulness of the details of Table 1, 
the main conclusions can be readily appreciated from visual 
inspection of the figures alone.

We will use Fig.~1 
\begin{figure}
\begin{center}
\resizebox{\sizefigs}{!}
{\includegraphics{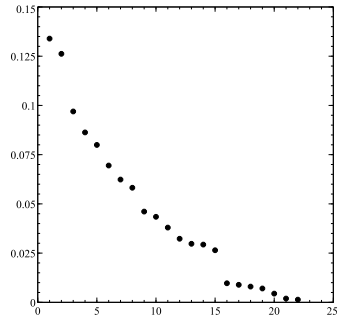}}
\caption{Fractional frequency as a function of rank for the phonemes
        of the reference case,
        defined here as the result of the basic phonemization (described in 
        Appendix B) on the data of the randomly chosen text.
From left to right, the 22 phonemes are: 
/e/,
/a/,
/o/,
/s/,
/i/,
/n/,
/\textfishhookr/,
/l/,
/d/,
/t/,
/k/,
/b/,
/u/,
/m/,
/p/,
/g/,
/x/,
/f/,
/r/,
/\textdyoghlig/,
/\textteshlig/,
/\textltailn/.
We use the International Phonetic Alphabet conventions.}
\end{center}
\end{figure}
as the reference for our analysis,
and compare the smoothness of the other figures 
in relation to the smoothness of this figure.
Fig.~1 represents the frequency distribution (fractional frequency versus rank)
for the chosen text after carrying out the basic phonemization scheme
described in Appendix B. The results are in agreement with
the literature (for a review, see Refs.~[18,19]). 
In this figure, the semiconsonant sounds [j] and [w]
(palatal and labiovelar approximants, respectively) 
are considered mere allophones of the vowels /i/ and /u/.
This case has a total of 22 phonemes.
As compared to Fig.~1, half of the remaining graphs have values of the smoothness 
which are considerably lower, while the other half of the graphs have 
comparable or greater values of the smoothness. 
The Shannon redundancy values for the four smoothest graphs (Figs.~1, 3, 4, 5)
are very similar among themselves and lower 
than the value for a 22-symbol 
Zipfian distribution, which is 15.9\%. The robustness of the redundancy value to changes of conditions 
in Figs.~1, 3, 4, 5  
may be an indication that the (naturally optimized) system sits close to 
a local minimum in the corresponding configuration space. 
 
Fig.~2 shows 
\begin{figure}
\begin{center}
\resizebox{\sizefigs}{!}
{\includegraphics{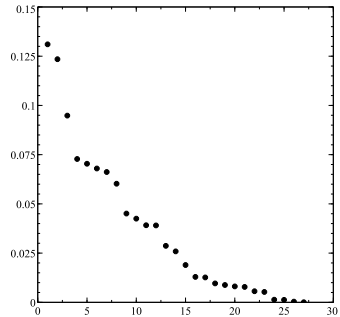}}
\caption{
Fractional frequency of the letters of the text under study. 
From left to right, the 27 letters are: 
e,
a,
o,
i,
r,
n,
s,
l,
d,
t,
c,
u,
m,
p,
b,
g,
v,
q,
h,
y,
f,
z,
j,
\~n,
x,
k,
\"u (acute accent marks have been ignored).
Letters make bad proxies for phonemes.}
\end{center}
\end{figure}
the distribution of letters ($N = 27$) for the same text,
without any phonemization. 
It can readily be appreciated by visual inspection that the corresponding 
distribution is considerably more abrupt. The computed value for the smoothness
is accordingly much lower.
Not surprisingly, letters make bad proxies for phonemes.

Fig.~3 
\begin{figure}
\begin{center}
\resizebox{\sizefigs}{!}
{\includegraphics{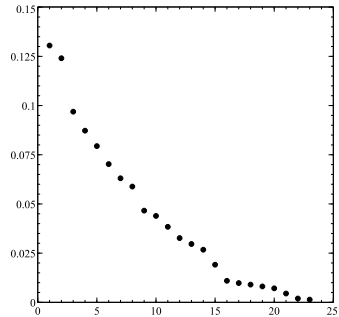}}
\caption{
Fractional phoneme frequency including the semiconsonant
/w/ as a separate 
phoneme from the vowel /u/.
From left to right, the 23 phonemes are: 
/e/,
/a/,
/o/,
/s/,
/i/,
/n/,
/\textfishhookr/,
/l/,
/d/,
/t/,
/k/,
/b/,
/m/,
/p/,
/u/,
/w/,
/g/,
/x/,
/f/,
/r/,
/\textdyoghlig/,
/\textteshlig/,
/\textltailn/. Note how /u/ and /w/, located respectively at positions
15 and 16 in this figure, remedy the gap after position 15 in Fig.~1.
}
\end{center}
\end{figure}
corresponds to the same conditions as those of Fig.~1 but making a 
distinction between /u/ and /w/ as separate phonemes (increasing thus 
the number of phonemes to 23).
Thus, the word ``su'' would be rendered /su/ while ``bueno'' would be rendered
/bweno/. In this last word, the two sounds /w/ and /e/ occur within
the same syllable (i.e. they form a diphthong), which is a condition for
the appearance of the semiconsonant phoneme. 
Instrumentally, we replaced all occurrences of the letter 
``u'' next to a vowel (on either side) by a ``w'', indicative of the separate 
phoneme /w/.  Non-diphthong adjacent occurrence of ``u'' and vowel 
(as in ``ba\'ul'') 
has a very low statistical frequency (probability $\sim 0.0001)$, and
therefore its neglect is inconsequential.
 
It is remarkable, and the main finding of this paper, 
that such a minor modification can smooth out the curve
so dramatically, remedying the gap after position 15 in Fig.~1.
This strongly suggests that the gap was an artifact of Spanish orthography,
where /u/ and /w/ are represented by a single letter (``u'').
This graph has the highest computed smoothness of all presented graphs.
If one takes this hint from complexity, 
we find thus good evidence that /w/ in Spanish be considered a 
separate phoneme. The gap between positions 2 and 3 in Figs.~1
and 3 remains somewhat intriguing. Non-smoothness at 
distribution extremes, however, is not uncommon 
in this type of plot, where the distribution couples to ``external conditions''  
(which in this case might correspond to morphological constraints, 
for example). Note also that
the gap after position 3 in Fig.~1 shows a much smaller {\it ratio} than the one
after position 15.

Fig.~4 
\begin{figure}
\begin{center}
\resizebox{\sizefigs}{!}
{\includegraphics{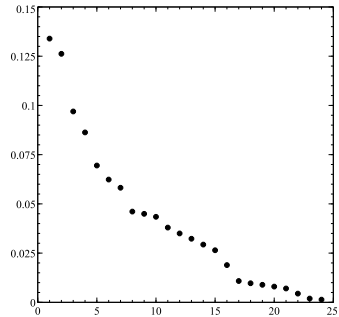}}
\caption{
Fractional phoneme frequency including both semiconsonants
/w/ and /j/ as a separate
phonemes from the respective vowels /u/ and /i/.
From left to right, the 24 phonemes are: 
/e/,
/a/,
/o/,
/s/,
/n/,
/\textfishhookr/,
/l/,
/d/,
/i/, 
/t/,
/k/,
/j/,
/b/,
/m/,
/p/,
/u/,
/w/,
/g/,
/x/,
/f/,
/r/,
/\textdyoghlig/,
/\textteshlig/,
/\textltailn/. 
}
\end{center}
\end{figure}
shows the result of treating /i/ and /j/ as separate phonemes
(in addition to /u/ and /w/). The results are at best inconclusive,
as the plot does not appear to show any improvement over Fig.~3.

Unlike the /w/ case, however, the /j/ case is problematic given that many
adjacent occurrences of ``i'' and vowel (either order) do not represent
true diphthongs so the substitution of /j/ for ``i'' is unwarranted. 
This inaccuracy can be remedied noting that Spanish spelling 
indicates the absence of a diphthong by a diacritical accent mark over 
the ``i'' (at least in careful speech).
This allows one to distinguish
/i/ versus /j/ minimal pairs (ignoring stress); contrast
``r\'{\i}o'' (/rio/) versus ``rio'' (/rjo/). 
The inclusion of 
diacritical mark considerations has the effect of
decreasing the number of instances of /j/ and increasing 
the number of instances of /i/.
Interestingly and reassuringly, 
such a swap results in a slightly smoother Fig.~5,
\begin{figure}
\begin{center}
\resizebox{\sizefigs}{!}
{\includegraphics{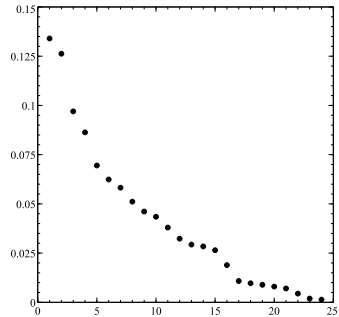}}
\caption{
Fractional phoneme frequency including both semiconsonants
/w/ and /j/ as a separate
phonemes from the respective vowels /u/ and /i/.
The difference between this figure and the previous one
is that this one includes the refinement of using
diacritical marks to better distinguish /i/ from /j/.
From left to right, the 24 phonemes are:  
/e/,
/a/,
/o/,
/s/,
/n/,
/\textfishhookr/,
/l/,
/i/, 
/d/,
/t/,
/k/,
/b/,
/m/,
/j/,
/p/,
/u/,
/w/,
/g/,
/x/,
/f/,
/r/,
/\textdyoghlig/,
/\textteshlig/,
/\textltailn/. 
}
\end{center}
\end{figure} 
as compared to Fig.~4
(the jump between positions 7 and 8 in Fig.~4 disappears).
We have thus evidence supporting the position that /j/ be also considered
a separate phoneme (although somewhat less strongly than in the case of /w/).  Fig.~5 shows a exponential decay for Spanish phonemes; such a distribution
appears in many complex networks such as electricity power grid transmission lines 
and airport traffic [5]. 
 
At this point, doubts might remain stemming from the fact that we have
based our analysis on a single text. 
To corroborate that the intuition was indeed correct, 
we repeated the procedure on ten additional texts, 
increasing thus the sample size by an order of magnitude. 
Reassuringly, the results were reproduced. 
Details can be found in Appendix C. 

On a separate final issue, and mainly for the sake of completeness, 
it is noteworthy 
that any attempt to consider diphthongs as separate phonemes,
a position not lacking distinguished supporters [20], 
leads to highly untenable results (Figs.~6 and 7), providing thus
very strong evidence against such a stand.
\begin{figure}
\begin{center}
\resizebox{\sizefigs}{!}
{\includegraphics{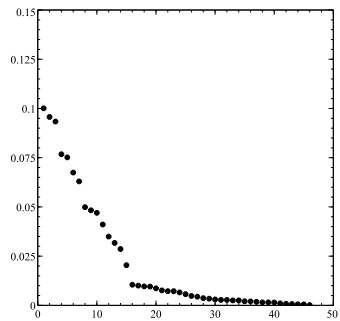}}
\caption{
Fractional phoneme frequency considering all diphthongs as separate phonemes.
A diphthong is thus defined for the sake of this figure as the confluence of 
{\it{any}} two vowels.
From left to right, the 46 putative phonemes are: 
/e/,
/a/,
/s/,
/o/,
/n/,
/\textfishhookr/,
/l/,
/d/,
/i/,
/t/,
/k/,
/b/,
/m/,
/p/,
/u/,
/g/,
/io/,
/x/,
/ia/, 
/f/,
/r/,
/ea/,
/ie/,
/oe/,
/ae/, 
/\textdyoghlig/,
/ee/,
/ai/,
/oa/,
/aa/,
/oi/,
/ue/,
/ei/,
/eu/,
/\textteshlig/,
/au/,
/ua/,
/ao/,
/eo/,
/\textltailn/,
/ou/,
/iu/,
/ui/,
/oo/,
/ii/,
/uo/. This figure and the next one provide strong evidence 
against the consideration of diphthongs as phonemes.}
\end{center}
\end{figure}
\begin{figure}
\begin{center}
\resizebox{\sizefigs}{!}
{\includegraphics{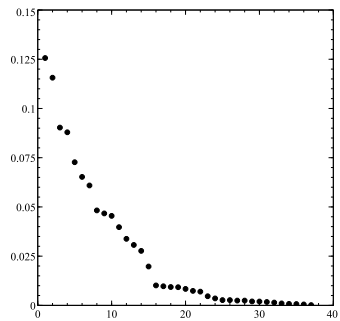}}
\caption{
Fractional phoneme frequency considering diphthongs as separate phonemes.
A diphthong is defined here as the confluence of two vowels such that at least
one of them is either /i/ or /u/.
From left to right, the 37 putative phonemes are: 
/e/,
/a/,
/s/,
/o/,
/n/,
/\textfishhookr/,
/l/,
/d/,
/i/,
/t/,
/k/,
/b/,
/m/,
/p/,
/u/,
/g/,
/io/,
/x/,
/ia/, 
/f/,
/r/,
/ie/,
/\textdyoghlig/,
/ai/,
/oi/,
/ue/,
/ei/,
/eu/,
/\textteshlig/,
/au/,
/ua/,
/\textltailn/,
/ou/,
/iu/,
/ui/,
/ii/,
/uo/.
The results do not change if one restricts further to 
the appearance of {\it unstressed} /i/ or /u/ as a criterion of diphthongness.}
\end{center}
\end{figure}

\section{Conclusions}
\label{}

Using simple ideas motivated by the behavior of complex systems
in general, and guided by physics principles, 
we have articulated
a straightforward criterion for phonemeness determination, 
an elusive problem
in linguistics.
The method consists of assessing 
relative frequency-distribution smoothness.

For the particular case we chose to study, we conclude that it is 
reasonable to 
consider the Spanish 
semiconsonants /w/ and /j/ as separate
phonemes from their respective vowel counterparts /u/ and /i/, 
although the evidence for the /j/ case is somewhat
less strong than for the /w/ case
and would benefit from further investigation.
(It is possible that the status of /j/ determination is complicated 
by the fact that its counting
involves dealing with the full consonant /\textdyoghlig/ as well.) 

As language has been so central a topic in the study of complexity, 
this discussion grants us, in addition, by setting the stage for establishing analogies with other physical systems, 
an opportunity to gain insight 
into emerging properties in the broader complex systems debate.

The present exploratory article has made the
tacit assumption, common in linguistics, that the
classification of the sounds of a language into distinct phonemes 
is a viable model. Some authors, however, warn us
that 
``contrast  must  be  treated  as  a  gradient  phenomenon  at 
the  phonological  level,  with  membership  of  a  phonemic  inventory  
being  a  matter  of degree'' [3].
Indeed, languages are known to make changes in their phonemic
inventory over centuries, and any state of transition 
would be expected to cause classification difficulties.
Perhaps a wider discussion is needed in this sense. 
The redundancy value for Fig.~3 is the highest among the smooth 
figures (Figs.~1, 3, 4, 5). We might speculate that the emergence of /w/ as
a separate phoneme might compensate for a loss of redundancy as
the Spanish phoneme inventory simplified over the centuries [21].

Future work could apply the methodology presented in this 
exploratory article to a variety of problematic phonological 
features throughout world languages.

The method could also be applied to study diachronic phoneme evolution
over the centuries, in order for example to measure phoneme phase
transitions. 

\newpage

\section*{References}
\label{}
\noindent
[1] C. Beckner, R. Blythe, J. Bybee, M.~H. Christiansen, W. Croft, N.~C. Ellis, 
J. Holland, J. Ke, D. Larsen-Freeman, T. Schoenemann,
Language is a complex adaptive system: Position paper,
Language Learning 59 (2009) 1--26.  

\noindent
[2] D. Larsen-Freeman, Complex Systems and Applied Linguistics, 
    Oxford University Press, 2008. 

\noindent
[3] J.~M. Scobbie, J. Stuart-Smith, Quasi-phonemic contrast and the fuzzy
    inventory: Examples from Scottish English, in: 
    Contrast in Phonology: Theory, Perception, Acquisition, 
    Mouton de Gruyter, 2008, 
    pp.~87--114. The quotation is from page 87.

\noindent
[4] R. Ferrer i Cancho, R. Sol\'e, The small world of human language,
Proc.~R. Soc.~Lond.~B 268 (2001) 2261--2265.

\noindent
[5] L. A. N. Amaral, A. Scala, M. Barth\'el\'emy, H. E. Stanley,
Classes of small-world networks, Proc. Natl. Acad. Sci. 97 (2000) 11149-11152.

\noindent
[6] S. T. Piantadosi, 
    Zipf's word frequency law in natural language: A critical review
    and future directions, Psychon Bull Rev. 21 (2014) 1112--1130. 

\noindent
[7] B. Sigurd, M. Eeg-Olofsson, J.~van de Weijer, Word length, sentence length
    and frequency--Zipf revisited, Studia Linguistica 58 (2004) 37--52.

\noindent
[8] Y. Tambovtsev, C. Martindale, Phoneme frequencies follow a Yule distribution,
    SKASE Journal of Theoretical Linguistics 4 (2007) 1--11.

\noindent
[9] B. Mandelbrot, Contribution a la th\'eorie math\'ematique des jeux de
communication, Publications de l'institut de statistique de l'universit\'e de
Paris 2 (1952) 1--124. 

\noindent
[10] G. K. Zipf, Human Behavior and the Principle of Least Effort: An
     Introduction to Human Ecology, Addison-Wesley, 1949.

\noindent
[11] S.~Thurner, R. Hanel, P. Klimek, Physics of evolution: Selection without fitness,
     Physica A 389 (2010) 747--753.

\noindent 
[12] J.~N.~Green, Spanish, in: The Romance Languages, Oxford University Press, 
     1988, pp.~79--130. 
 
\noindent
[13] V.~Mart\'{\i}nez-Paricio, The intricate connection between diphthongs 
     and stress in Spanish, Nordlyd 40 (2013) 166--195. 

\noindent
[14] U. Schild, A. B. C. Becker, C. K. Friedrich, Processing of syllable stress is functionally different from phoneme processing and does not profit from literacy acquisition, Frontiers in Psychology 5 (2014) article 530. 

\noindent
[15] J.~I.~Hualde, Quasi-phonemic contrasts in Spanish, 
     in: WCCFL 23: Proc.~of the 23rd West Coast Conference on Formal     
     Linguistics, 2004, pp.~374--398.

\noindent
[16] R.~J.~Hyndman, G. Athanasopoulos, Forecasting: Principles and Practice,
     OTexts, 2014.

\noindent
[17] C. E. Shannon, A mathematical theory of communication, 
     Bell System Technical Journal 27 (1948) 379--423, 623--656.

\noindent
[18] M.~Guirao, M.~A. Garc\'{\i}a Jurado, Frequency of occurence of phonemes in 
     American Spanish,
     Revue qu\'eb\'ecoise de linguistique 19 (1990) 135--149.

\noindent
[19] H.~E.~P\'erez, Frecuencia de fonemas, 
     e-rthabla 1 (2003). 

\noindent
[20] T.~Navarro Tom\'as, Escala de frecuencia de los fonemas espa\~noles, 
     in: Estudios de Fonolog\'{\i}a Espa\~nola, 
     Syracuse University Press, 1946, pp.~15--30.

\noindent
[21] R. Penny, A History of the Spanish Language,
    Cambridge University Press, 2002.

\noindent
[22] J. L. Borges, El Aleph, Emec\'e, 1957.


\appendix

\section{Assumptions}
\label{}

We list here explicitly all the assumptions made during the analysis.
The assumptions are not weak; we rely on 
past successes of low-resolution approaches to complex systems.
 
First, we assumed there is nothing peculiar about the chosen text,
the 1945 short story {\it El Aleph} by J. L. Borges [22]
(which was selected randomly from a list of Spanish language literature), 
and that its length is adequate.
The properties of language under study are thus assumed to be robust enough
that they reproduce themselves
at the level of a single text.  This issue was corroborated by subsequent considerations, as 
described in appendix C. 
 
Second, we worked under the assumption that 
text is a good enough proxy for speech. 
One of the difficulties arising from this assumption is that
there exist artifacts of spelling, e.g. the 
use of two letters for the same phoneme (``b'' and ``v'');
this difficulty is partly countered by our simple phonemization scheme
presented in Appendix B. 
The other related difficulty is that, even if transcription were perfectly phonological,
the way a person writes is not the way he or she speaks, so 
our measurement is only approximate in this sense
(and the distance if even greater with
spontaneous, non-careful speech).

Third, we used for the analysis the features of typical Latin American Spanish
(which means no distinction between /s/ and /\texttheta/, on one hand,
and also typically between /\textdyoghlig/ and /\textturny/, on the other),
as this type of Spanish corresponds to a majority of speakers of the language.
 
Fourth, it was assumed that /\textdyoghlig/ and /j/ remain 
separate phonemes.

Finally, as is usual in phonological analyses,
the stress phoneme was ignored, so words 
differing only in their stress, 
such as ``p\'ublico'' and ``public\'o,'' would be 
indistinguishable in our analysis.  
(An exception to this assumption was made for
the analysis leading to Fig.~5, 
as discussed in Section 3.)

\section{Phonemization used in case study}
\label{}
The following is the basic phonemization [12] process performed
on the raw data.
Even though this procedure is crude,
going beyond this scheme would be unproductive, 
given the low resolution stance of the paper as a whole.

Once the text was lower-cased, diacritical marks were eliminated
(``\~n'' was maintained, however, as it represents the phoneme /\textltailn/). 
We then performed the following replacements
(indicated by arrows) as a rough approximation to a 1-to-1 mapping between
symbols and phonemes. The underscore sign stands for an empty space.  
\begin{enumerate}
   \item qu $\rightarrow$ k
   \item v $\rightarrow$ b
   \item x $\rightarrow$ ks  
   \item z $\rightarrow$ s
   \item ch $\rightarrow$ v 
   \item h $\rightarrow$ \_
   \item ce $\rightarrow$ se 
   \item ci $\rightarrow$ si
   \item c $\rightarrow$ k
   \item j $\rightarrow$ x
   \item ge $\rightarrow$ xe
   \item gi $\rightarrow$ xi 
   \item gue $\rightarrow$ ge
   \item gui $\rightarrow$ gi
   \item y\_ $\rightarrow$ i\_
   \item y, $\rightarrow$ i,
   \item y. $\rightarrow$ i.
   \item ll $\rightarrow$ y
   \item \_r $\rightarrow$ \_q
   \item rr $\rightarrow$ q
   \item sr $\rightarrow$ sq
   \item nr $\rightarrow$ nq
   \item lr $\rightarrow$ lq
\end{enumerate}
For the analysis leading to Fig.~5, 
the diacritical mark over the ``i'' was 
maintained in order to discriminate between /i/ and /j/.
In this case, rules 8, 11, 13 were complemented with
c\'{\i} $\rightarrow$ s\'{\i},
g\'{\i} $\rightarrow$ x\'{\i},
gu\'{\i} $\rightarrow$ g\'{\i}.

We note that, after the aforementioned substitutions, 
all of the final symbols in the working text are identical
to the International Phonetic Alphabet symbols, except for the following:
``v'' stands for /\textteshlig/, ``y'' stands for /\textdyoghlig/, 
``q'' stands for /r/, ``r'' stands for /\textfishhookr/,
``\~n'' stands for /\textltailn/. We make no use 
of /\texttheta/ or /\textturny/ for the reasons explained in Appendix A.

\section{Replication of results on additional texts}
\label{}

In order to check the ideas of the present paper, we performed
the procedure on ten additional texts 
of Spanish language literature (1940-1960 period), 
increasing thus the sample size by 
an order of magnitude to a total of 202 thousand phonemes. 

The results are shown in Figs.~C.1, C.2 and C.3, 
which have the same conditions as Figs.~1, 3 and 4, respectively,
but use the aggregate data. 
Fig.~C.1 shows the jump after position 15, which disappears once /w/ 
is considered as a separate phoneme from /u/, in Fig.~C.2.
Fig.~C.3 shows the inclusion of /j/ as separate from /i/ as well, 
and also distinctly evidences smoothing.

It is quite remarkable that, even though
the texts varied in length, date, 
country of origin and writer gender, 
{\it every single one of them showed the jump after position 15} in
their (not shown) C.1 plots, even as phoneme order varied from 
text to text.    
Although the situation was not as clearcut for the 
/j/ case (which is closer to the distribution tail), the
results were nevertheless there as well. 
In Spanish, /w/, /u/, /j/ and /i/ are separate phonemes.  

\setcounter{figure}{0} 
\begin{figure}
\begin{center}
\resizebox{\sizefigs}{!}
{\includegraphics{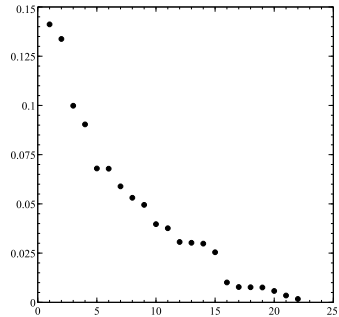}}
\caption{Aggregate-data fractional frequency as a function of 
        phoneme rank, after the basic phonemization (described in 
        Appendix B) has been carried out. 
        The conditions are the same as for Fig.~1.     
From left to right, the 22 phonemes are: 
/a/,
/e/,
/o/,
/s/,
/n/,
/i/,
/\textfishhookr/,
/l/,
/d/,
/t/,
/k/,
/b/,
/u/,
/m/,
/p/,
/g/,
/x/,
/r/,
/\textdyoghlig/,
/f/,
/\textteshlig/,
/\textltailn/.}
\end{center}
\end{figure}

\begin{figure}
\begin{center}
\resizebox{\sizefigs}{!}
{\includegraphics{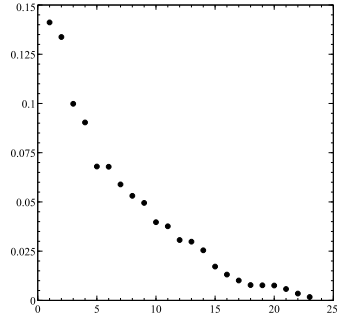}}
\caption{Aggregate-data fractional frequency as a function of 
        phoneme rank, including the semiconsonant /w/ as a separate phoneme from 
        the vowel /u/. The conditions are thus the same as for Fig.~3. 
From left to right, the 23 phonemes are: 
/a/,
/e/,
/o/,
/s/,
/n/,
/i/,
/\textfishhookr/,
/l/,
/d/,
/t/,
/k/,
/b/,
/m/,
/p/,
/u/,
/w/,
/g/,
/x/,
/r/,
/\textdyoghlig/,
/f/,
/\textteshlig/,
/\textltailn/.}
\end{center}
\end{figure}

\begin{figure}
\begin{center}
\resizebox{\sizefigs}{!}
{\includegraphics{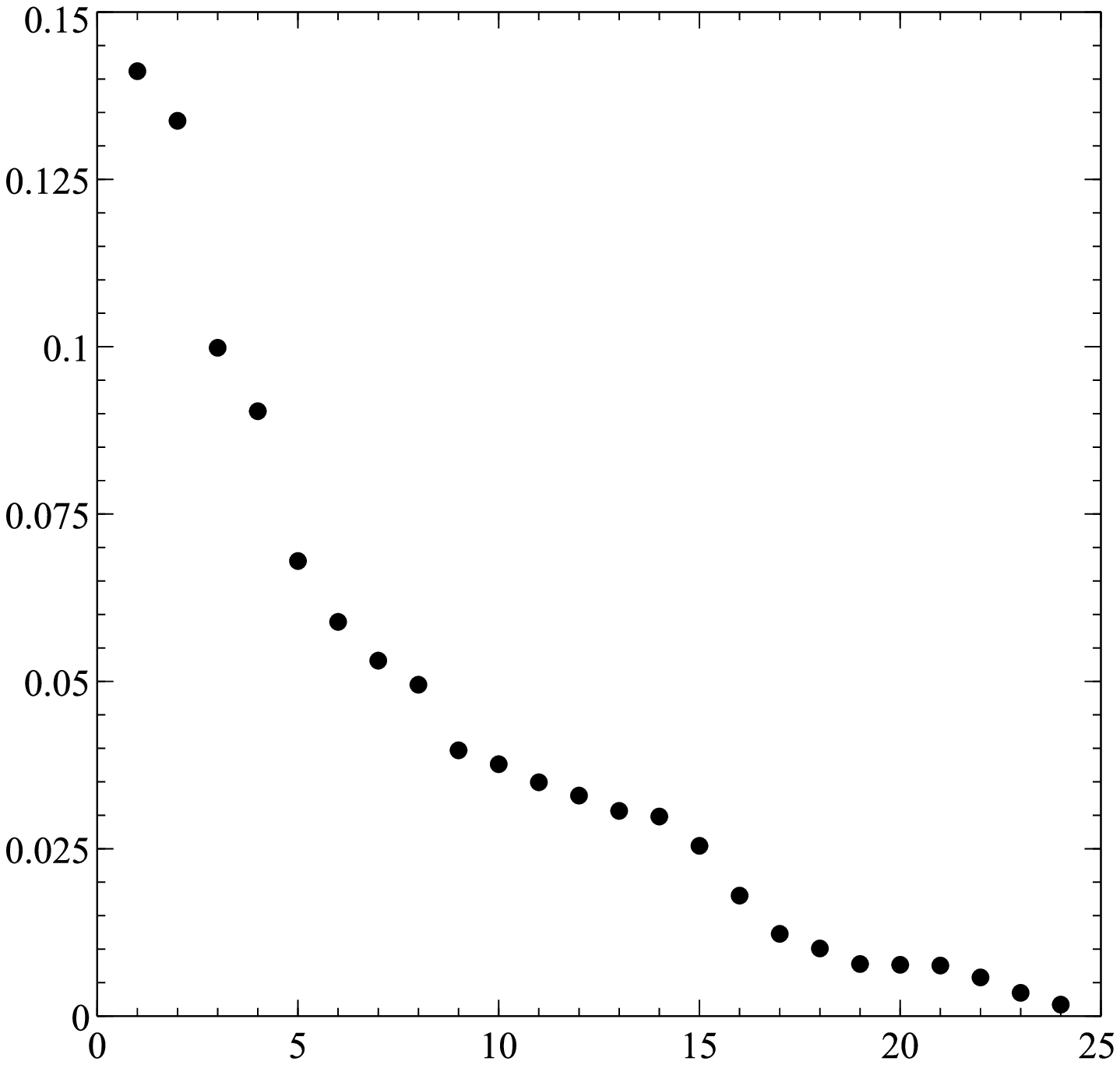}}
\caption{Aggregate-data fractional frequency as a function of 
        phoneme rank, including the semiconsonants /w/ and /j/ 
        as a separate phonemes from the respective vowels /u/ and /i/.
        the vowel /u/. The conditions are thus the same as for Fig.~4. 
From left to right, the 24 phonemes are: 
/a/,
/e/,
/o/,
/s/,
/n/,
/\textfishhookr/,
/l/,
/d/,
/t/,
/k/,
/i/,
/j/,
/b/,
/m/,
/p/,
/u/,
/w/,
/g/,
/x/,
/r/,
/\textdyoghlig/,
/f/, 
/\textteshlig/,
/\textltailn/.}
\end{center}
\end{figure}



\end{document}